%% file: main_elsarticle-template-num.tex
\documentclass[preprint,12pt]{elsarticle}

\usepackage{amssymb}
\usepackage{amsmath}
\usepackage{algorithm}
\usepackage{algorithmic}
\usepackage{multirow}
\usepackage{url}
\usepackage{hyperref}
\usepackage{xcolor}
\usepackage[symbol]{footmisc}

\journal{Nuclear Physics B}

\begin{document}

\begin{frontmatter}



\title{PREVENT: Proactive Risk Evaluation and Vigilant Execution of Tasks for Mobile Robotic Chemists using Multi-Modal Behavior Trees}

\author[UoL]{Satheeshkumar Veeramani\corref{cor1}}
\ead{satheeshkumar.veeramani@liverpool.ac.uk}
\author[UoL]{Zhengxue Zhou}
\author[UoL]{Francisco Munguia-Galeano}
\author[UoL]{Hatem Fakhruldeen}
\author[UoR]{Thomas Roddelkopf}
\author[UoR]{Mohammed Faeik Ruzaij Al-Okby}
\author[UoR]{Kerstin Thurow}
\author[UoL]{Andrew Ian Cooper\corref{cor1}}
\ead{aicooper@liverpool.ac.uk}

\cortext[cor1]{Corresponding authors.}

\affiliation[UoL]{
    organization={Department of Chemistry and Material Innovation Factory, University of Liverpool},
    city={Liverpool},
    country={United Kingdom}
}
\affiliation[UoR]{
    organization={Center for Life Science Automation, University of Rostock},
    city={Rostock},
    country={Germany}
}

\begin{abstract}
Mobile robotic chemists are a fast growing trend in the field of chemistry and materials research. However, so far these mobile robots lack workflow awareness skills. This poses the risk that even a small anomaly, such as an improperly capped sample vial could disrupt the entire workflow. This wastes time, and resources, and could pose risks to human researchers, such as exposure to toxic materials. Existing perception mechanisms can be used to predict anomalies but they often generate excessive false positives. This may halt workflow execution unnecessarily, requiring researchers to intervene and to resume the workflow when no problem actually exists, negating the benefits of autonomous operation. To address this problem, we propose PREVENT a system comprising navigation and manipulation skills based on a multimodal Behavior Tree (BT) approach that can be integrated into existing software architectures with minimal modifications. Our approach involves a hierarchical perception mechanism that exploits AI techniques and sensory feedback through Dexterous Vision and Navigational Vision cameras and an IoT gas sensor module for execution-related decision-making. Experimental evaluations show that the proposed approach is comparatively efficient and completely avoids both false negatives and false positives when tested in simulated risk scenarios within our robotic chemistry workflow. The results also show that the proposed multi-modal perception skills achieved deployment accuracies that were higher than the average of the corresponding uni-modal skills, both for navigation and for manipulation.
\end{abstract}


\begin{highlights}
\item Novel multi-modal behavior-tree system for detecting workflow anomalies for mobile robotic chemists.
\item Combines AI-driven vision and olfactory modalities to detect and interpret safety hazards.
\item We designed and conducted experiments to prove the robustness and reliability of the system.
\item Robot safely performed both navigation and manipulation tasks in a powder X-ray diffraction workflow.
\end{highlights}

\begin{keyword}
Robotics and Automation in science \sep mobile manipulator robots \sep Multimodal perception.


\end{keyword}

\end{frontmatter}



\input{introduction}

\input{Methodology}
\input{results}
\input{Conclusion}

\section*{Acknowledgements}
{AIC thanks the Royal Society for a Research Professorship (RSRP\textbackslash S2\allowbreak\textbackslash232003). This work was supported by the Engineering and Physical Sciences Research Council (EPSRC) under grant agreements EP/\allowbreak V026887/1 and EP/\allowbreak Y028759/1, the Leverhulme Trust through the Leverhulme Research Centre for Functional Materials Design, and the H2020 ERC Synergy Grant Autonomous Discovery of Advanced Materials under grant agreement no. 856405.}

\section*{Data availability}
The code, models, and data can be accessed at \url{https://github.com/satheezv/PREVENT.git}. Videos of the experiments for all tasks are available at \protect\url{https://youtu.be/B07r0J6y8OQ}












\bibliographystyle{elsarticle-num}
\bibliography{references}
\end{document}

%% file: introduction.tex
\section{Introduction and Related Work}

\subsection{Mobile Robotic Chemists}
The deployment of automated or semi-autonomous mobile robotic chemists (MRCs) in chemistry and materials research laboratories is a rapidly emerging trend, both in academic and industrial settings. MRCs can accelerate the research process by using standard laboratory equipment, much like a human researcher, to perform tasks such as solid/liquid handling,  synthesis (\textit{e.g.,} using Chemspeed ISynth platforms), and analysis (\textit{e.g.,} liquid chromatography–mass spectrometry) \cite{burger2020mobile,lunt2024modular}. Most set-ups use a centralised orchestration software to coordinate the workflow equipment, including robots, and to communicate with them via available communication frameworks, such as the Robot Operating System (ROS) or ZeroMQ \cite{dai2024autonomous,fakhruldeen2022archemist}. This orchestration software issues commands to MRCs only when necessary, allowing the robots to handle routine maintenance tasks —such as charging and joint sensor calibration— during idle periods. This approach enables the robot to work for relatively long periods (\textit{e.g.}, 8 continuous days in \cite{burger2020mobile}) with relatively few interruptions, thus greatly improving experimental throughput with respect to traditional manual experiments.  In large part, this is because such robots can be programmed to work 24/7 and to perform highly repetitive tasks without error.  This can greatly outpace human experimentation, even if individual unit operations, such as moving around the laboratory, are slower, which is often the case.

\begin{figure}[h]
    \centering
    \includegraphics[width=\linewidth,trim=0.05cm 0cm 0cm 0cm,clip]{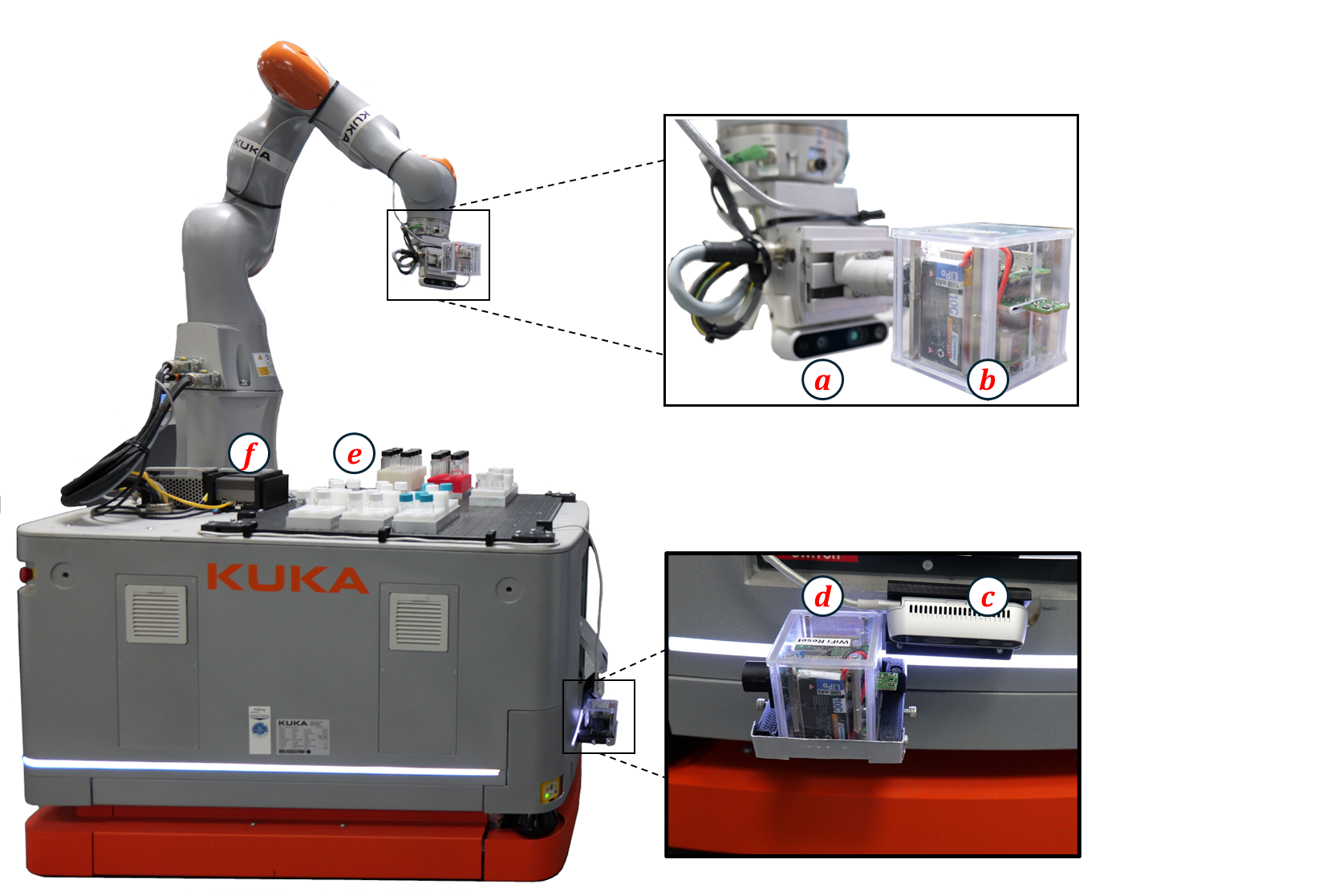}
    \caption{%
        \textbf{Mobile Robotic Chemist} equipped with vision and olfactory perception modalities. 
        Components \textcircled{\scriptsize\textcolor{red}{\textbf{a}}} and \textcircled{\scriptsize\textcolor{red}{\textbf{b}}} denote the vision and olfactory sensory modules of the robot arm, respectively. 
        Components \textcircled{\scriptsize\textcolor{red}{\textbf{c}}} and \textcircled{\scriptsize\textcolor{red}{\textbf{d}}} correspond to the vision and olfactory sensors of the mobile base. 
        Components \textcircled{\scriptsize\textcolor{red}{\textbf{e}}} and \textcircled{\scriptsize\textcolor{red}{\textbf{f}}} represent the robot's deck and onboard computer, respectively.%
    }
    \label{fig:MRC}
\end{figure}

However, it is impossible to engineer robotic solutions that are totally foolproof, and MRCs generally assume that the workflow is proceeding according to plan. That is, there is little feedback to indicate when a mistake has occurred, such as the robot accidentally dropping a sample vial. Hence, MRCs tend to execute assigned tasks in an open-loop fashion, with the exception of standard cobot features such as laser scanning (to avoid collisions when traversing the laboratory) and force-feedback, which can indicate that the robot arm or manipulated object has encountered an unanticipated obstacle  \cite{burger2020mobile}. The creation of solutions to the problem of unexpected workflow errors could determine the prospects for large-scale deployment of MRCs in the future. Laboratories, while complex, present a semi-structured environment, and there are significant opportunities for training MRCs to understand hazard scenarios and to act accordingly.

\subsection{Integrating Behavior Trees into Workflow Orchestration Architectures}


Behavior Trees (BTs) present a compelling solution in the context of automated robotic laboratory workflows, where safety, robustness, and adaptability are paramount. Their modular design allows complex procedures to be broken down into reusable and interpretable behavior units, enhancing explainability and maintainability. Furthermore, their inherent reactivity makes them especially suitable for managing hazardous tasks and responding to dynamic environmental changes, ensuring both operational flexibility and safety~\cite{ghzouli2023, akkaladevi2025, martin2025towards}. The hierarchical and modular structure of BTs enables intuitive task decomposition and supports dynamic reconfiguration at runtime, allowing robots to adapt to changing task conditions and recover from unexpected failures~\cite{akkaladevi2025}. 
Moreover, they offer greater readability and extendability compared to FSM or raw code~\cite{Suddrey2022}. As a result, BTs have been  used in many different robotic application domains, including multi-robot systems~\cite{Best2024}, human-robot interaction~\cite{Scherf2023}, autonomous driving~\cite{kato2018} and industrial assembly~\cite{akkaladevi2024}. 


Structurally, BTs are directed acyclic graphs composed of a root node, control-flow (non-leaf) nodes, and execution (leaf) nodes. Execution begins at the root, which periodically ticks its child nodes and traverses the tree based on the semantics of control-flow nodes. Each node returns one of three statuses: running (in progress), success (task completed), or failure (task unsuccessful). Condition nodes, which evaluate specific predicates, return only success or failure. Core control-flow nodes include \emph{fallback}, \emph{sequence}, and \emph{parallel} nodes. A fallback node succeeds if any of its children succeed and fails only if all fail; a sequence node fails if any child fails and succeeds only if all succeed; a parallel node ticks all children concurrently and succeeds only when all children report success. This combination of control structures provides a powerful abstraction for building complex, reactive, and maintainable robotic behaviors~\cite{Tanaka2023}.


\subsection{Hazards in self-driving labs}


Chemical hazards such as accidental spillages, broken glassware, and contaminated accessories, such as gloves, can occur in laboratories for various reasons, such as equipment malfunction, improper object handling, or human error.  If left unattended, these hazards can lead to dangerous situations such as exposure to toxic chemicals, fire, or explosions, which have caused fatalities in the past \cite{leggett2012identifying}  \cite{menard2020review}. Moreover, mobile robots use Li-ion batteries that can undergo thermal runaway and explosions if exposed to fire~\cite{kong2018li}. Self-driving laboratories (SDLs) \cite{cooper2025lira} involve a wide range of equipment, such as mobile robot arms \cite{burger2020mobile}, mobile robots \cite{liu2012floyd, liu2013mobile}, fixed robot arms \cite{jiang2023autonomous}, and gantry devices \cite{dai2024autonomous,lunt2024modular} that transfer, dispense and analyze chemicals, and handle different types of labware. Today, there are relatively few methods to allow robots to determine whether it is safe to proceed an assigned task. For example, the presence of sensors such as LIDAR (Light Detection and Ranging) and F/T (Force-Torque), do not allow robots to detect small but chemically hazardous obstacles—such as vials, contaminated gloves, or spillages—during navigation. Likewise, failure scenarios during manipulation, such as dropping a vial, encountering a missing object, or attempting to grasp a rack with a misplaced vial, may go unnoticed and could cause safety issues or workflow failure.   

\subsection{Modalities and perception for robots in self-driving labs}

Human chemists are trained to spot hazards and to handle them by adhering to defined safety protocols. However, it is challenging to transfer these skills to robots. In general, robots use various external sensors such as vision, depth, and, in the context of laboratories, measurements of volatile organic compound (VOC) concentrations. This involves internal sensors such as touch, force/torque, and joint states modalities to gather feedback and to take precise actions \cite{munguia2023affordance, song2024vlm}. This sensing can be carried out in a unimodal fashion—that is, using a single sensory modality—or in a multi-modal fashion by combining multiple sensing modalities, depending on the task complexity and its requirements \cite{munguia2025chemist, pizzuto2024accelerating, song2024multiagent}. For example, Al et al.~\cite{al2024ambient} integrated an ambient sensing modality with a mobile robot to read VOC, carbon dioxide, and temperature levels around a robot, with the aim of detecting leakage of gases and chemical vapors. When a serious chemical hazard occurs, the VOC level in the environment rises, and VOC sensors can identify such conditions. To enable robots to perceive and monitor ongoing chemistry experiments, Dharvish et al.~\cite{DARVISH2025101897} integrated multiple modalities, including a pH meter, potentiostat–analyzer, and lab scales.

Our previous works have shown that binding AI models with multi-model perception and information-fusion at the model-level helps robots to make better decisions with increased flexibility \cite{fakhruldeen2025multimodal, Zhou2025VLM}. For instance, in~\cite{cooper2025lira}, we implemented an edge-computing system for MRCs, creating an onboard pipeline by integrating HandEye cameras, software frameworks, and AI models to enable localization, inspection, and reasoning processes. In another relevant work, Pizzuto et al.~\cite{pizzuto2024accelerating} coupled vision and F/T modalities with a Deep Reinforcement Learning technique and developed a policy for a robotic arm to execute a sample scraping task, which also showed that using several sources of data improves the learning capabilities of robots, even for complex tasks. Vision modality combined with Convolutional Neural Networks (CNNs) was also used to detect material solubility \cite{pizzuto2022solis}.

\subsection{VLMs for hazard detection}
Another approach in the literature that integrates multiple sources of information is Contrastive Language–Image Pre-training (CLIP) \cite{radford2021learning}. CLIP is a Neural Network-based model that learns visual concepts with natural language supervision (\textit{i.e.}, image and text tokens), which allows zero-shot capabilities. This allows us to implement classifiers, where it is necessary to design labels (\textit{i.e.}, text elements) to provide the model with contextual information. However, these designed labels heavily influence both the model prediction and biases. To design classes, a clear list of all potential objects is required; otherwise, CLIP is not very efficient \cite{dorbala2022clip, ZHANG2025105210}. In consequence, to predict hazards in a robotic workflow, it is recommended to use CLIP to classify identified anomalies rather than detecting anomalies \cite{muttaqien2025attention, gao2024physically}.  A better approach to address this issue is to use CNN models, which perform better when it comes to binary prediction \cite{pizzuto2022solis}. 

\subsection{Contribution}
To overcome the identified feedback and decision making gaps for MRCs, we propose a novel multimodal hazard-aware framework that enables proactive and context-sensitive responses during autonomous chemistry operations. Specifically, proposed PREVENT system incorporates: \textcircled{\scriptsize 1} two new skills that fuse vision and olfactory modalities to perceive the environment, either continuously (during navigation) or on-demand (prior to arm interaction with station objects), thereby enabling real-time hazard detection and classification; \textcircled{\scriptsize 2} a new decision-making mechanism to take appropriate actions or countermeasures upon hazard detection; and \textcircled{\scriptsize 3} bidirectional communication interfaces to keep both MRCs (Fig.~\ref{fig:MRC}) and human chemists in the loop. 

Unlike existing systems that rely solely on unimodal perception or predefined safety zones, our approach uniquely integrates multimodal (vision + olfactory) sensing with adaptive decision-making, enabling instant anomaly recognition and human-aware feedback through CNN and VLM models. We further conducted rigorous experiments on a mobile robotic chemist performing solid-state workflow tasks to evaluate the robustness and reliability of the proposed system under real laboratory conditions.

%% file: Methodology.tex
\section{Methodology}

\begin{figure*}[h!]
\vspace{-15pt}
    \centering
    \includegraphics[width=\linewidth]{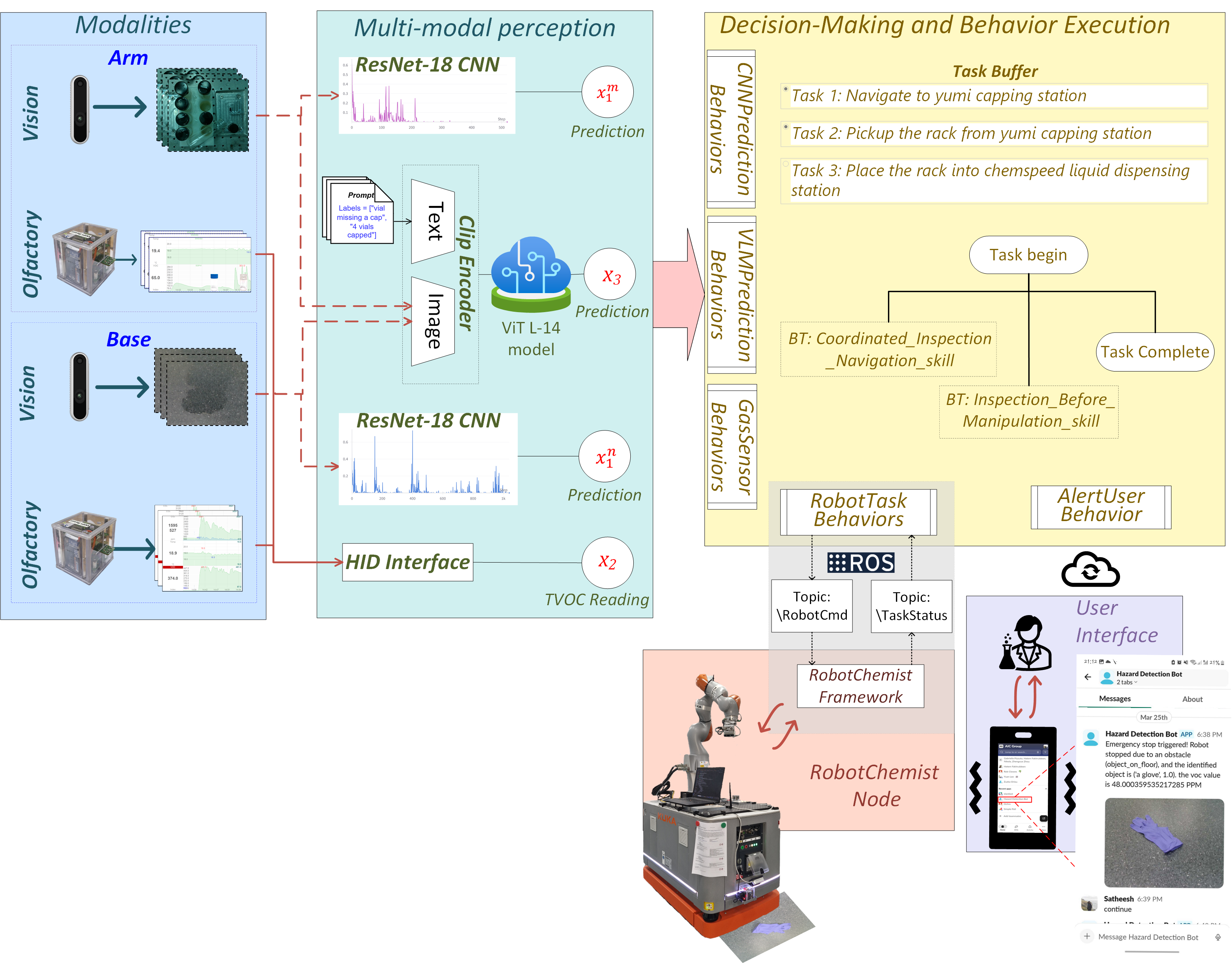}
    \caption{\textbf{PREVENT system architecture}
    that enables hazard mitigation by integrating perception, decision-making, execution, and human and robot interface modules.}
    \label{fig: system}
\end{figure*}

This section presents a hazard mitigation strategy that equips MRCs with enhanced navigation and manipulation skills, designed to detect and respond to workflow hazards in SDLs. It also describes the system architecture (Fig.~\ref{fig: system}) that implements the strategy, outlines its constituent modules, and elaborates on the implementation of the navigation and manipulation skills.

\subsection{Proposed Strategy}
We propose to solve the problem of mitigation of workflow hazards of MRCs by taking high-level decisions based on the relevant decision mechanism, defined as \textit{skills}, based on multi-modal inputs and computed through the following equation. 

\begin{equation}
a^* =  f(a_i \mid x_1, x_2, x_3)
\label{eq:general}
\end{equation}

Here, the best decision for the robot to take \( a^* \) is selected based on the input modalities \( x_1, x_2, x_3 \) where $a_i$ is one of the three possible decisions the system can take: $a_1$ proceed with the task, $a_2$ halt and resume automatically after a secondary check, and $a_3$ halt and wait for a user command to continue. Let \( x_1 \in \{0, 1\} \) be the binary predictions of the CNN, where \( 0\) denotes \texttt{no hazard} and \(1\) denotes \texttt{hazard}. The VOC index readings from the olfactory modality \footnote{The IoT olfactory sensor module was developed at Center for Life Science Automation (CELISCA) at the University of Rostock, Germany.} are given by \( x_2 \in \mathbb{Z} \), and \( x_3 \in \mathcal{L} = \{ l_1, l_2, \ldots, l_j  \mid l_j\in[0,1]\} \) are categorical predictions of the different modalities (\textit{e.g.}, spillage, capping failure or no problem detected), respectively. More specifically, \( x_1\) and \( x_3\) are the predictions of vision-based modalities, whereas \( x_2\) corresponds to atmospheric VOC levels measured by the olfactory modality. 

Aiming to simplify this high-level problem, we further decompose it into two sub-problems: navigation and manipulation skills. The navigation skill is denoted as:

\begin{equation}
a^{n*} = f_n(a_i \mid x_1^n, x_2^n, x_3^n)
\label{eq:general_nav}
\end{equation}

Where the decision function \( f_n \) determines the best navigation action \(a^{n*}\). The manipulation skill is given by:

\begin{equation}
a^{m*} = f_n(a_i \mid x_1^m, x_2^m, x_3^m)
\label{eq:general_man}
\end{equation}

Where the decision function \( f_m \) determines the best manipulation action \(a^{m*}\). Therefore, each of the preceding functions is responsible for calculating the respective best actions for the robot based on inputs from its respective modalities. The core idea behind the derivation of these equations is that the robot requires continuous hazard perception during its navigation, whereas it only needs to observe the interaction frame once to detect potential hazards that obstruct its respective manipulation task. 

\subsection{System Architecture and Implementation}

To implement the hazard mitigation strategy introduced in the previous subsection, we designed the system shown in Fig.~\ref{fig: system}, which comprises modules ranging from perception to execution, allowing the MRC to perform the proposed skills. These modules include the multi-modal perception and prediction, decision making and execution, and the interface modules. This system can be integrated easily into existing workflow automation software infrastructures, particularly between the orchestration software and the robot interfaces.

\subsubsection{Skills and behaviors}

Here, a skill refers to the logical combination of task execution, hazard perception, and decision-making behaviors with in the PREVENT system. Two such skills are developed to address the identified gaps by handling workflow anomalies during navigation and manipulation tasks, respectively.  Both skills use two vision modalities—based on CNN and VLM models, respectively—for binary presence detection and detailed hazard classification, along with an olfactory modality to detect VOC levels in the surrounding environment.

\begin{figure*}[t]
    \centering
    \includegraphics[width=\linewidth]{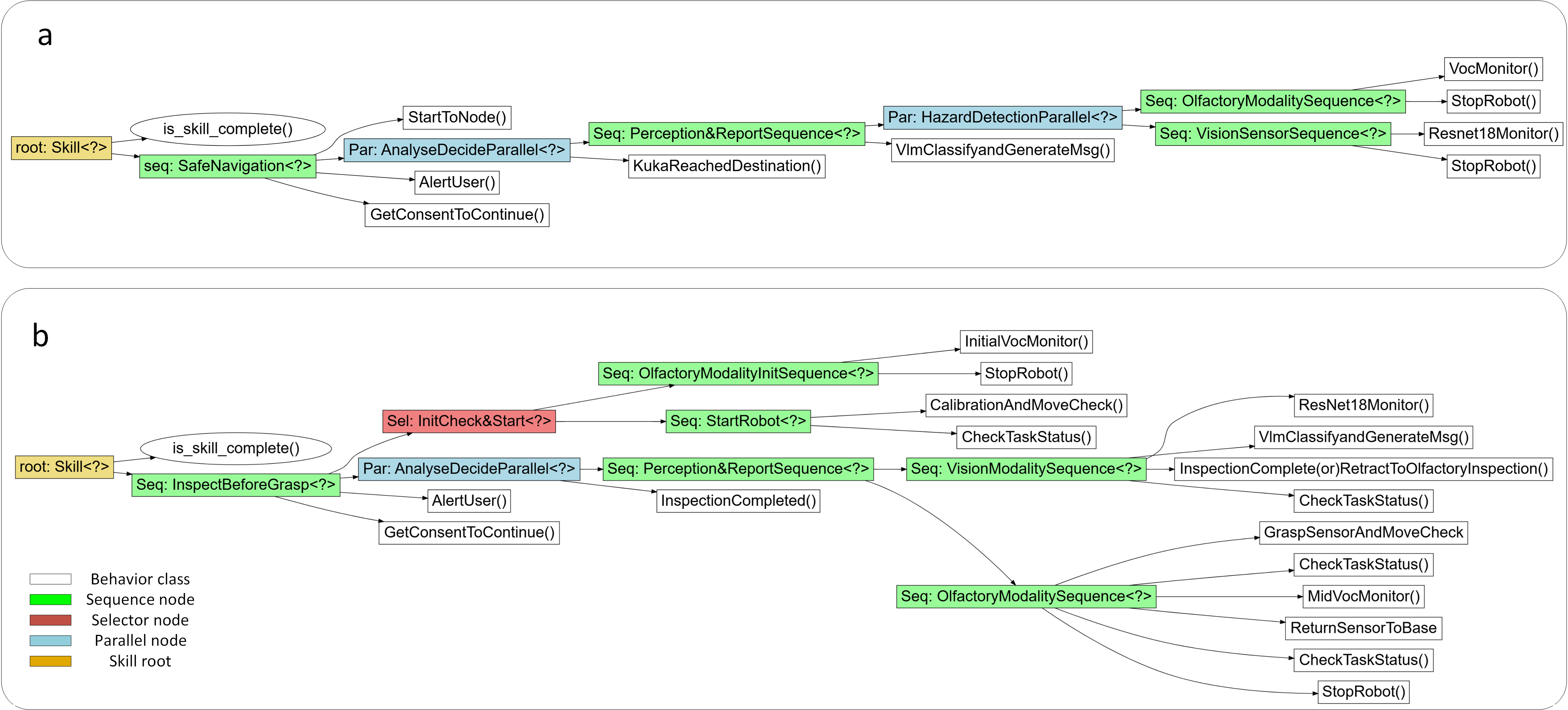}
    \caption{Behavior tree structures of both skills: \textbf{(a)} CIN skill and \textbf{(b)} IBM skill.}
    \label{fig:skills}
\end{figure*} 

\paragraph{Coordinated Inspection Navigation (CIN) skill}

During navigation, the MRC typically moves between two stations to transfer labware, or to a docking point for charging. The monitoring process in this context must be real-time and continuous to identify potential hazards in the path. To support this, the proposed CIN skill continuously monitors the floor ahead of the robot using vision (within 0.7 m in front of the robot) and olfactory modalities (approximately 1 m surrounding the robot). The behavior tree is designed based on the execution logic of the CIN skill as presented in algorithm \ref{alg:navigation}.

\begin{algorithm}[h]
\fontsize{8pt}{10pt}\selectfont
\caption{CIN skill}
\label{alg:navigation}
\renewcommand{\algorithmicendif}{\vspace{-1.25em}}
\renewcommand{\algorithmicendloop}{\vspace{-1.25em}}
\renewcommand{\algorithmicendwhile}{\vspace{-1.25em}}
\begin{algorithmic}
\REQUIRE \textit{dest\_node}, \textit{robot\_task\_id}, $\mathcal{L}$, $\mathcal{L}_{\text{safe}}$, $\mathcal{L}_{\text{unsafe}}$, $x_2^n$, $T_{safe}$, \textit{user\_id}.
\STATE start navigation to \textit{dest\_node}
\WHILE{not \textit{dest\_node}}
    \STATE read $x_2^n$, capture $image$ and predict $x_1^n$
    \IF{$x_2^n > T_{safe}$ \textbf{or} $x_1^n == hazard$}
        \STATE stop the robot
        \STATE predict $x_3^n$
        \IF{$x_3^n \in \mathcal{L}_{\text{unsafe}}$}
            \STATE notify user: $\{x_1^n, x_2^n, x_3^n\}$ + $image$
            \STATE request user consent
            \WHILE{not $response == continue$ }
                \STATE wait for $response$
            \ENDWHILE
        \ENDIF
    \ENDIF
\ENDWHILE
\end{algorithmic}
\end{algorithm}

Once the system receives a navigation task assigned to the robot from the orchestration framework, it initiates the CIN skill with the behavior \textit{StartToNode()}, which sends the task command to the robot. It then activates the \textit{ResNet18Monitor()} and \textit{VocMonitor()} behaviors to continuously monitor the navigation path. If either behavior detects a hazard, the \textit{StopRobot()} behavior is triggered. Subsequently, the \textit{VlmClassifyandGenerateMsg()} behavior classifies the detected hazard and decides whether to continue navigation—if the predicted hazard is labeled as $\mathcal{L}_{\text{safe}}$—or to pause and notify the user through \textit{AlertUser()} behavior if it is labeled as $\mathcal{L}_{\text{unsafe}}$. Finally, the \textit{GetConsentToContinue()} behavior makes the robot wait until it receives the \texttt{continue} command from the user and then resumes the task. The sequence, selector, and parallel composite nodes were used to build the logic as shown in the fig. \ref{fig:skills}a.

\paragraph{Inspection Before Manipulation (IBM) skill}
When performing manipulation tasks, the MRC interacts with stations primarily to pick up or place racks, individual sample vials, or other labware. Unlike navigation, this process does not require continuous monitoring, but targeted inspection of a specific interaction is essential. To achieve this, the proposed IBM skill moves the robot to a predefined \textit{check\_pose}, from where both the vision and olfactory modalities can perceive the target location. A behavior tree structure is designed based on the execution logic of the IBM skill, is presented in the algorithm \ref{alg:manipulation} and fig. \ref{fig:skills}b.

\begin{algorithm}[h]
\fontsize{8pt}{10pt}\selectfont
\caption{IBM skill}
\label{alg:manipulation}
\renewcommand{\algorithmicendif}{\vspace{-1.25em}}
\renewcommand{\algorithmicendloop}{\vspace{-1.25em}}
\renewcommand{\algorithmicendwhile}{\vspace{-1.25em}}
\begin{algorithmic}
\REQUIRE \textit{assigned\_task}, \textit{robot\_task\_id}, $\mathcal{L}$, $\mathcal{L}_{\text{safe}}$, $\mathcal{L}_{\text{unsafe}}$, $x_2^m$, $T_{safe}$, \textit{user\_id}. 
\STATE Start skill and read $\{x_2^m\}$
\IF{$x_2^m < T_{safe}$}
    \STATE Move to $check\_pose$ and predict $\{x_1^m\}$
    \IF{$x_1^m = hazard$}
        \STATE predict $\mathcal{L}$
        \IF{$\mathcal{L} == \mathcal{L}_{\text{unsafe}}$}
            \STATE Mid\_Olfactory\_sensing\_operations()
            \STATE notify \textit{user}: $\{x_1^m, x_2^m, x_3^m\}$ + $image$, and request $response$
            \WHILE{not $response == continue$ }
                \STATE wait for $response$
            \ENDWHILE
        \ENDIF
    \ENDIF
\ENDIF
\end{algorithmic}
\end{algorithm}

The system initiates the IBM skill with the \textit{InitialVocMonitor()} behavior to check for chemical hazards before executing the \textit{CalibrationAndMoveCheck()} behavior. After \textit{check\_pose}, the skill executes the vision \textit{ResNet18Monitor()} behavior for binary hazard prediction. If the behavior returns \texttt{no hazard}, the IBM skill terminates the inspection and sends the task execution command to the MRC. However, if the behavior returns \texttt{hazard}, the skill conducts secondary checks using the \textit{VlmClassifyandGenerateMsg()} and \textit{MidVocMonitor()} behaviors to classify the hazard and assess the VOC level at the station. Based on the results of the secondary check, the skill decides whether to continue execution or to trigger the \textit{AlertUser()} behavior. Finally, the \textit{GetConsentToContinue()} behavior makes the robot wait until it receives the \texttt{continue} command from the user, after which it resumes the task.

\subsubsection{Utilities and interfaces}
The PREVENT system contains various utility classes to interface with the visual and olfactory modalities, CNN and VLM models, and APIs (Application Programming Interfaces) such as \texttt{RobotChemist} and \texttt{slack\_sdk} for communication with the robot and the human chemist, respectively.

\paragraph{Modalities}
The system is currently equipped with two vision and two olfactory modalities, with one of each assigned to the CIN and IBM skills. The vision modality uses support image acquisition and includes a Laplacian-based blurriness check to ensure image quality. The CNN and VLM models for the vision modalities are designed to support model loading, training, validation, and real-time hazard prediction. The olfactory modality utility class provides functionalities for reading environmental data, including temperature, TVOC (Total Volatile Organic Compounds), IAQ (Index of Air Quality), CO\textsubscript{2}, NO\textsubscript{x}, and H\textsubscript{2} levels.

\begin{figure*}[h]
    \centering
    \includegraphics[width=0.75\linewidth]{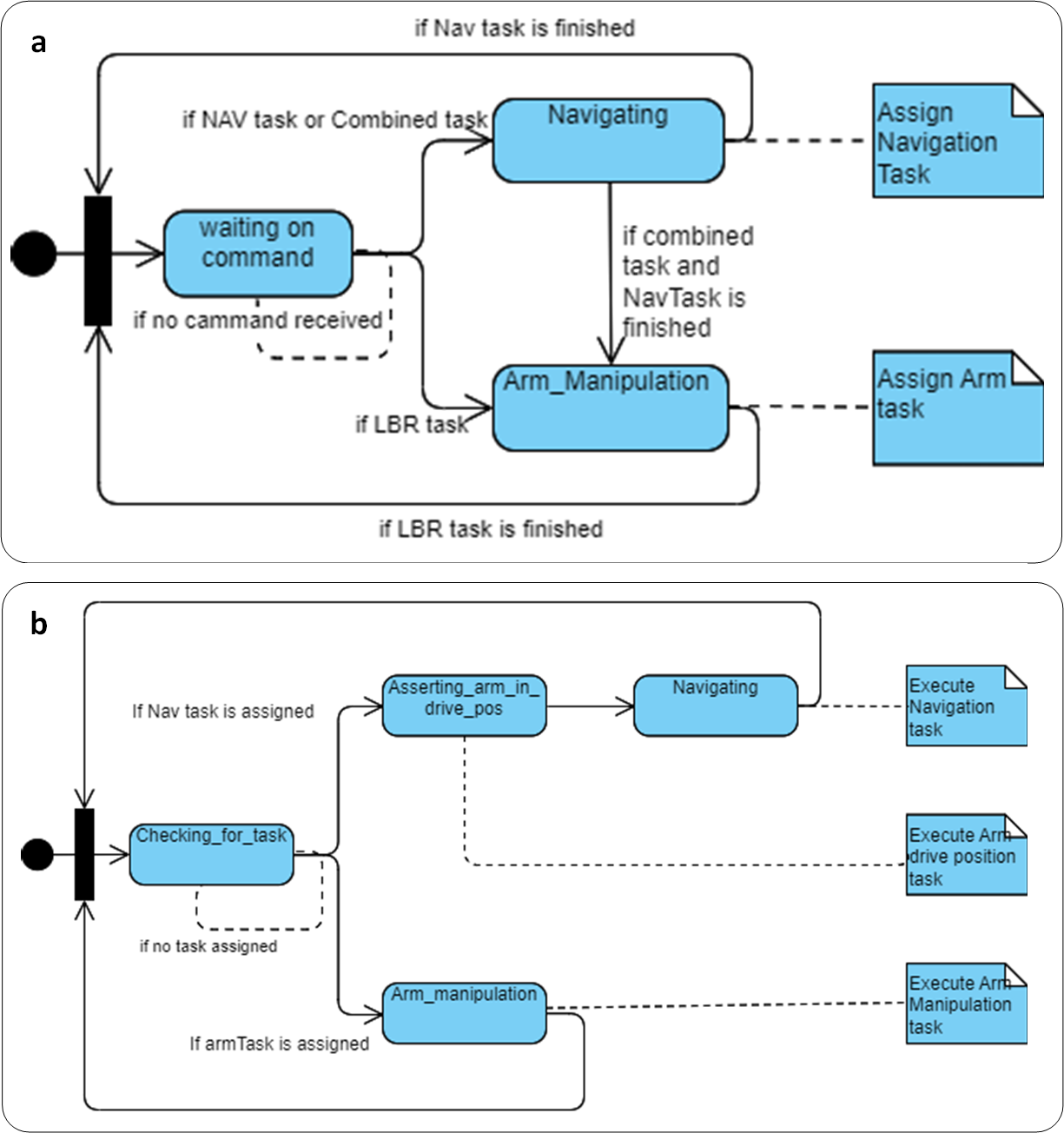}
    \caption{Custom-made driver of the MRC showing \textbf{(a)} task assignment and \textbf{(b)} task execution.}
    \label{fig:driver}
\end{figure*}

\paragraph{APIs}
The robot was controlled through KUKA \texttt{Sunrise.OS} software using a custom made driver (fig.\ref{fig:driver}). This API receives tasks through a ROS task message that contains information about the task type, task name, and location. If the task type is either manipulation \textit{(LBR task)} or navigation \textit{(NAV task)}, the execution is straightforward: the system performs the task while the corresponding skill (CIN or IBM) monitors the execution throughout. However, if the system receives a \textit{combined\_task}, it first executes the CIN skill by sending a navigation message with the target location. Upon receiving this message, the API initiates the navigation operation. Once the CIN skill completes successfully, the IBM skill is then triggered, and the robot performs the corresponding manipulation task upon receiving the relevant task details. This API also updates the status of task execution and the MRC back to the system via ROS messages.

%% file: results.tex
\section{Experimental Evaluation \&  Discussion}

\begin{figure*}[h!]
    \centering
    \includegraphics[width=\linewidth]{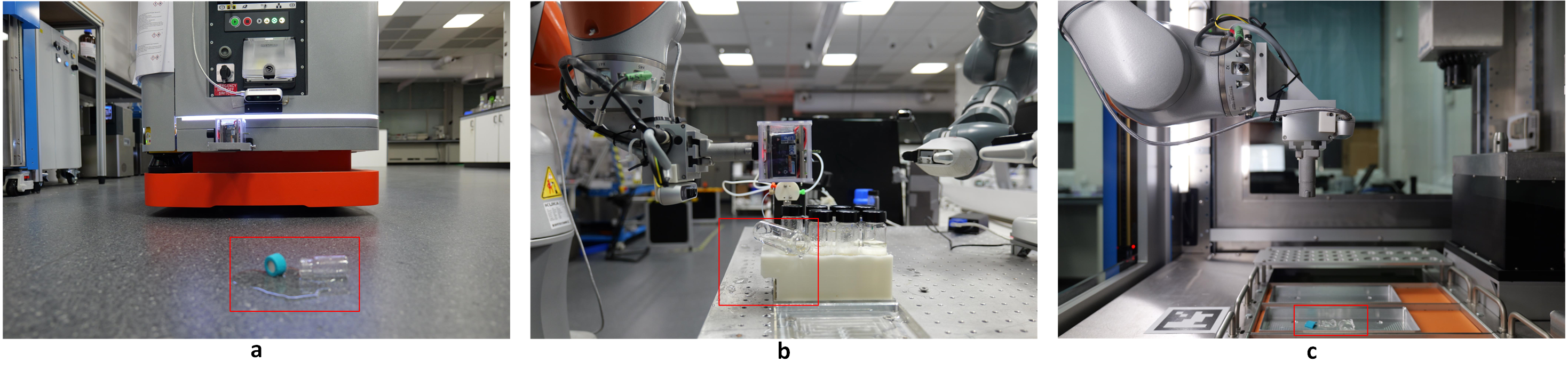}
    \caption{Photographs showing simulated hazards in the MRC workflow; \textbf{(a)}\ Vial\ \&\ chemical\ spillage\ during\ task\ T1, leading to a vial and chemicals on the lab floor that is too small to be detected by LIDAR,\ \textbf{(b)} \ Improper\ vial\ drop\ in\ capping\ station\ during\ task\ T2, \textbf{(c)}\ Broken\ glass\ pieces\ in\ the Chemspeed\ station\ during\ task\ T3. The relevant hazard is highlighted by a red rectangle in each case. }
    \label{fig:tasks}
    \vspace{-5pt}
\end{figure*}

This section presents the details of the implementation and our experimental results. First, we selected one of our most complex MRC workflows to date—powder X-ray diffraction (PXRD) of solid-state materials—which involves a diverse range of manipulation tasks and interactions with multiple types of equipment. This choice was designed to test our PREVENT system in a challenging and realistic research environment. Three robot tasks from the workflow that present possible hazards were identified (Fig. \ref{fig:tasks}), and experiments were conducted to: 1) identify optimal behavior sequences to avoid false negatives and false positives; 2) evaluate the test and deployment accuracies of each modality individually across all three tasks, and; 3) assess robustness and reliability of both the skills.    \footnote{Videos of the experiments for all tasks are available at \protect\url{https://youtu.be/B07r0J6y8OQ}}

The experimental setup consists of a KUKA KMR-iiwa (a mobile robotic manipulator) equipped with two Intel RealSense D435i RGB-D cameras, as shown in  Fig.~\ref{fig:MRC}, one mounted on the gripper and the other on the mobile base. Two detachable gas sensor modules \cite{al2024ambient} are also attached—one at the front of the base and the other on the deck. An onboard computer with and 16 GB RAM, which runs both CIN and IBM skills, is also mounted on the deck. 
The capping station includes an ABB-YUMI robot with a rack containing 8 slots for holding 50 ml vials, and the liquid dispensing station comprising a Chemspeed-FLEX platform composed of a gantry robot with three rack holders. To test the Olfactory modality, we used solvents such as ethyl alcohol (C$_2$H$_6$O), acetone (CH$_3$COCH$_3$), and isopropanol (C$_3$H$_8$O) as representative but low-hazard examples, noting that the VOC sensor is broad spectrum and responds to all VOC's irrespective of toxicity. These liquids are also flammable.

\subsection{Experiments for performance and robustness evaluation}

\subsubsection{Data collection, training and finetuning}
To fine-tune the CNN and VLM models, a total of 5,003 images were collected using both the dexterous and navigation vision cameras across all three tasks (Fig. \ref{fig:tasks}). These images were used for training (CNN only), fine-tuning (VLM only), and validation. For data collection, a front-mounted navigation camera captured a 70 cm floor area for navigation tasks, while the hand-eye camera was used for manipulation tasks (see Fig. \ref{fig:predictions}). An automated script was used to stochastically command the MRC base to visit all nodes multiple times for navigation data collection. For manipulation tasks, the robot arm was moved to a predefined \textit{check\_pose} to capture images of the target locations.

The datasets collected across the three task types are task scenario-centric, meaning each scene is labeled as either safe or hazardous based on the overall environmental condition, rather than on specific object-level details. As discussed earlier, the CNN models are used for binary classification (\texttt{hazard} or \texttt{no hazard}) to detect even small anomalies within a scene. Therefore, the fine-tuning process emphasizes learning the characteristics of the \texttt{no hazard} class, so that any reasonable deviation from this baseline is classified as a \texttt{hazard} scene. To train in navigation task T1, the no hazard class included clean and unobstructed floor conditions, while the hazard class involved potential laboratory hazards such as contaminated gloves, chemical spillages, vials, broken glass, and tools placed along the robot’s path. For the pickup task T2 at the capping station, the no hazard class featured a complete vial rack with eight capped vials containing chemicals. In contrast, the hazard class involved anomalies such as missing vials, uncapped vials, knocked-over vials, chemical spillages, and obstructions blocking the grasp frame. In the drop task T3 at the Chemspeed station, the no hazard class represented a clean, empty tray, while the hazard class included anomalies such as broken glass pieces or chemical spillages present on the tray. 

\subsubsection{Evaluating prediction stability across modalities}

We conducted stability tests for all modalities in each task to assess their individual deployment accuracy in detecting both hazardous and safe scenarios (Fig. \ref{fig:predictions}). We created scenarios by manually placing hazards such as (low hazard) chemical spillages, contaminated gloves, and other lab objects on the floor to evaluate the CIN skill. For the IBM skill, scenarios included few uncapped vials in a rack, a knocked-over vial with chemical spillage, and the presence of obstructing objects, such as broken pieces of glass. For the safe scenarios, we retained the default workflow scenarios without modification.

We tested three different visual modalities for each task (30 runs for $T_1$ and 50 runs for $T_2$ and $T_3$): one ResNet-18-based model (fine-tuned) for binary classification of \texttt{hazard} and \texttt{no hazard}, and two ViT L/14-based model (both zero-shot and fine-tuned) for the classification of identified hazard types ($\mathcal{L}_{\text{safe}}$, $\mathcal{L}_{\text{unsafe}}$). As shown in Table \ref{tab:modality_accuracy}, the ResNet-18 model demonstrated high accuracy, achieving over 96\% in tasks $T_1$ and $T_2$, and 92\% in $T_3$. It was effective in identifying anomalies along the robot’s path. Most failed predictions occurred due to small optically transparent obstructions in $T_1$ and $T_3$, as well as an uncapped vial placed in the slot of the rack farthest from the camera frame in $T_2$. The zero-shot ViT-L/14 model demonstrates limited accuracy when handling unseen image-text pairs. Consequently, its deployment accuracy on this model across the three tasks is relatively low—63\%, 40\%, and 48\%, respectively. 
In contrast, the fine-tuned ViT-L/14 model significantly improves performance, achieving 90.41\%, 90.20\%, and 98.00\% accuracy on Tasks 1, 2, and 3, respectively. The primary failure mode in Task 1 arises from the model misclassifying irregular floor textures as foreign objects (7/73). For Task 2, most misclassifications occur when both obstruction and spillage are present in the same image—a scenario not represented in the training set (5/51). In Task 3, the rare failure cases stem from instances where only a very small portion of the vial appears at the edge of the image, making detection particularly challenging (1/50).



For the olfactory modality, the safety threshold ($T_{\text{safe}}$) for VOC levels was derived based on the Eq.~\ref{eq:threshold} as 2.5 PPM, and each run was initiated only when the VOC value was below this threshold. We note here that there is no single safe VOC threshold—it depends on the chemical substances being used—but this was chosen as a representative threshold for chemical substances with low to moderate safety risks. For task $T_1$, the robot was programmed to navigate to a random location. When the modality detected a VOC level above the threshold, the robot halted its motion and the value was recorded. The deployment accuracy was 88\%, as the robot failed to halt immediately in three instances when a hazard was suddenly introduced. 

\begin{table}[h!]
\centering
\caption{Modality performance across navigation and manipulation skills.}
\label{tab:modality_accuracy}
\begin{tabular}{c|c|c|c|c|c}
\textbf{Skill} & \textbf{Task} & \textbf{Modality} & \textbf{Model} & \textbf{Test} & \textbf{Deployment} \\
\textbf{} & \textbf{} & \textbf{} & \textbf{} & \textbf{Accuracy} & \textbf{Accuracy} \\
\hline
\multirow{4}{*}{CIN} 
    & \multirow{4}{*}{T1} & Visual & ResNet-18 \textbf{(FT)} & 99.10\% & 96.7\% \\
    &                     & Visual & ViT-L/14 \textbf{(ZS)}  & -- & 63\% \\
    &                     & Visual & ViT-L/14 \textbf{(FT)}  & 97.52\% & 90.41\% \\
    &                     & Olfactory & -- & -- & 88\% \\
\hline
\multirow{4}{*}{IBM} 
    & \multirow{4}{*}{T2} & Visual & ResNet-18 \textbf{(FT)} & 99.05\% & 96\% \\
    &                     & Visual & ViT-L/14 \textbf{(ZS)}  & -- & 40\% \\
    &                     & Visual & ViT-L/14 \textbf{(FT)}  & 100\% & 90.20\% \\
    &                     & Olfactory & -- & -- & 90\% \\
\hline
\multirow{4}{*}{IBM} 
    & \multirow{4}{*}{T3} & Visual & ResNet-18 \textbf{(FT)} & 96.01\% & 92\% \\
    &                     & Visual & ViT-L/14 \textbf{(ZS)}  & -- & 48\% \\
    &                     & Visual & ViT-L/14 \textbf{(FT)}  & 100\% & 98\% \\
    &                     & Olfactory & -- & -- & 90\% \\
\end{tabular}
\vspace{2pt}
\footnotesize{
\textit{Notes:} 
T1 – Navigation to capping station; 
T2 – Pickup rack from capping station; 
T3 – Place rack at Chemspeed station. 
\textbf{FT} – Fine-tuned model; 
\textbf{ZS} – Zero-shot model.
}
\end{table}

\begin{figure}[h!]
    \centering
    \includegraphics[width=\linewidth]{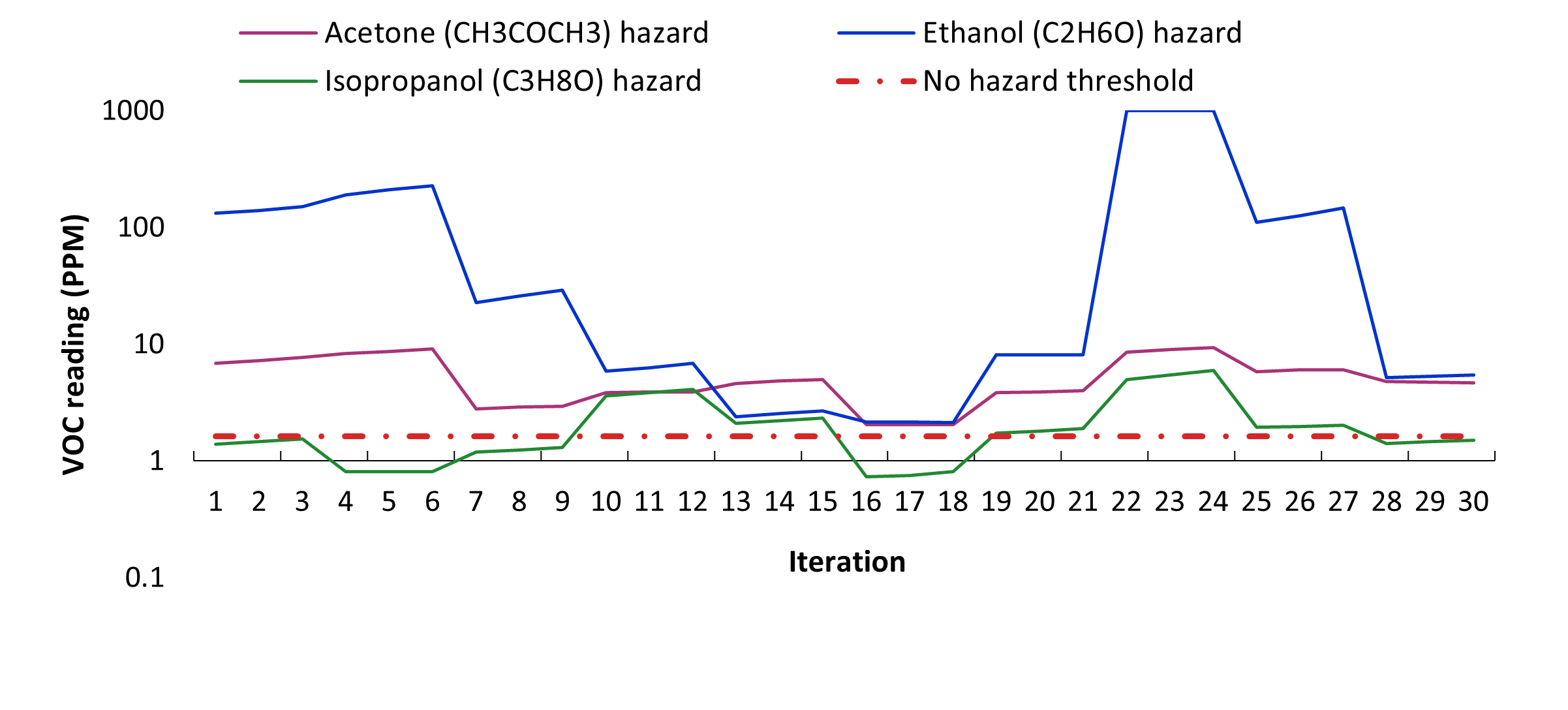}
    \caption{Olfactory modality - VOC values during various workflow Hazards.}
    \label{fig:VOC_plot}
\end{figure}

For tasks $T_2$ and $T_3$, the robot was programmed to pick up the sensor from its deck, move to the \textit{check\_pose}, and record the VOC level. Fig.~\ref{fig:VOC_plot} shows the results of stability testing for the olfactory modality, with 30 iterations each for all three solvents under the \texttt{hazard} condition, and 30 iterations for the \texttt{no hazard} condition using sealed vials. The VOC readings of sealed vials vary for each chemical due to several factors, such as volatility and ambient temperature. From the figure, it can be observed that the \texttt{hazard} readings for isopropanol frequently fall below $T_{\text{safe}}$, and the remaining readings are also close to this threshold. These results indicate that the olfactory modality alone may not be sufficient for effective chemical hazard detection in all cases. This limitation is also the primary reason for the reduction in the accuracy of the olfactory modality, as shown in Table~\ref{tab:modality_accuracy}. Based on the results, we also infer that the modality detects actual spillages more effectively than chemicals present in unsealed vials, which is understandable since the chemical is partially contained (and less hazardous) in the latter case. The deployment accuracy in these tasks was 90\%, with the other causes of failure being partially closed vials and uncapped vials located in the farthest slots of the rack, especially when containing highly volatile chemicals such as acetone due to their faster evaporation.

\begin{equation}
\text{T}_{\text{safe}} = \frac{1}{ct} \sum_{i=1}^{t} \left( S_{\text{ACE}, i} + S_{\text{EtOH}, i} + S_{\text{IPA}, i} \right)
\label{eq:threshold}
\end{equation}

Where $t$ is the total number of trials, c is the total number of solvents, $i$ is the iteration, $S_{\text{ACE}, i}$ is the VOC value of sealed Acetone, $S_{\text{EtOH}, i}$ is the VOC value of sealed Ethanol, and $S_{\text{IPA}, i} $ is VOC value for sealed Isopropanol.

\begin{figure}[t]
\vspace{-20pt}
    \centering
    \includegraphics[width=\linewidth]{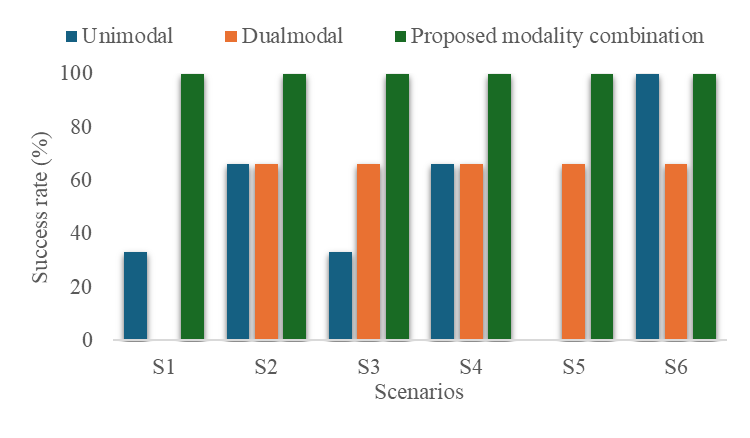}
    \caption{Skill execution with different combination of modalities under potential false positive and false negative conditions.}
    \footnotesize \textit{Scenarios:} S1 – chemical spillage away from the robot path, S2 – Chemical spillage on robot path covered by a paper, S3 - Sudden appearance of contaminated glove in front of the robot (to check the reaction time), S4 - spillage away from the target object, S5 - object obstruction the grasp position, S6 - improperly capped vial in the target rack.
    \label{fig:tasks_plot}
\end{figure}

\subsubsection{Evaluating modality combinations to avoid false negatives and false positives}

\begin{figure}[t]
    \centering
    \includegraphics[width=\linewidth]{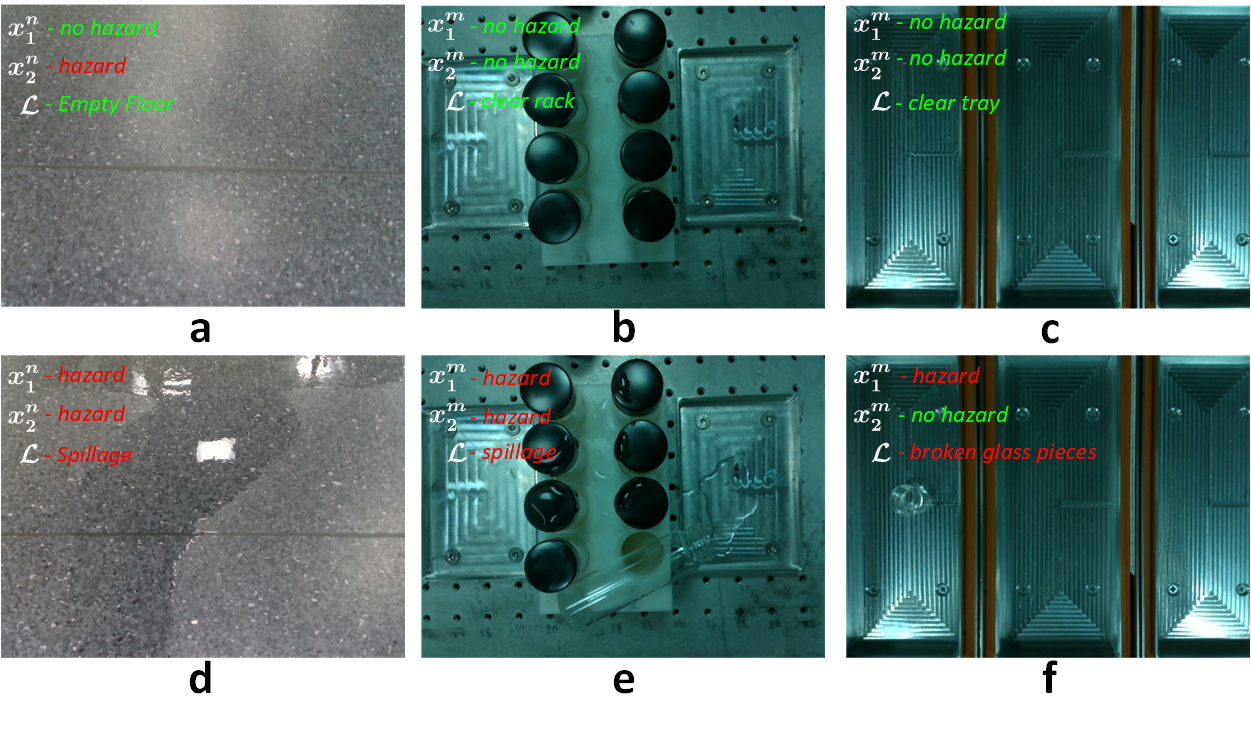}

    \caption{Hazard perception by MRC. In \textbf{(a)} and \textbf{(d)}, the CIN skill showtakes actions $a_2^n$ and $a_3^n$, respectively, based on predictions from base modalities. Similarly, IBM skill takes actions $a_1^m$ and $a_3^m$ in \textbf{(b)} and \textbf{(e)}, and $a_1^m$ and $a_2^m$ in \textbf{(c)} and \textbf{(f)}, respectively, based on predictions from arm modalities.}
    \label{fig:predictions}
    \vspace{-5pt}
\end{figure}

A total of 54 experimental runs were conducted initially to determine the optimal sequence of perception behaviors for both skills. First, we defined the set of high-level actions for both skills ($a_i^n$ and $a_i^m$, where $i \in \{1, 2, 3\}$) and tested the modalities individually (uni-modal) and in different combinations (dual-modal and proposed multi-modal) on various scenarios. The results in Fig.~\ref{fig:tasks_plot} show that, when deployed individually, the modalities fail mainly by making false negatives in scenarios S2, S3 and S5. Dual-modal combinations performs comparatively well but make false positives and stops execution, creating false alerts and requests for human intervention, even if it is safe to continue the execution. These results also demonstrate that the proposed modality combination for both the skills performs well across all scenarios and are highly effective in avoiding both false positives and false negatives (Fig.~\ref{fig:predictions}) .

\subsubsection{Testing skills' robustness and reliability}
To evaluate the robustness of the proposed skills, experiments were conducted on all three tasks (Fig.~\ref{fig:tasks}). A total of 60 runs were performed—20 for each skill. For each skill, 10 runs were executed under \texttt{no hazard} conditions to evaluate false positives, and 10 runs under \texttt{hazard} conditions to evaluate false negatives. To compare execution times, 60 runs of No Skill Executions (NSE) were also conducted. The results in Table \ref{tab:end2end_tasks_test} show that both skills have completed all runs without any false positives or false negatives.

\begin{table}[h!]
\centering
\caption{Task performance under different hazard types.}
\label{tab:end2end_tasks_test}
\begin{tabular}{c|c|c|c|c|c}
\textbf{Task} & \textbf{Skill} & \textbf{Runs} & \textbf{Hazard} & \textbf{Avg. Task} & \textbf{Success} \\
              &                &               & \textbf{Type}   & \textbf{Duration (s)} & \textbf{Rate (\%)} \\
\hline
\multirow{6}{*}{T1} 
 & \multirow{3}{*}{CIN} & \multirow{3}{*}{20} & NH  & 128.2 $\pm$ 0.2   & \multirow{3}{*}{100} \\
 &                      &                     & OH  & 179.2 $\pm$ 3.9   & \\
 &                      &                     & LSH & 460.7 $\pm$ 104.5 & \\
\cline{2-6}
 & \multirow{3}{*}{NSE} & \multirow{3}{*}{20} & NH  & 119.1 $\pm$ 0.6   & \multirow{3}{*}{50} \\
 &                      &                     & OH  & \textit{fail}     & \\
 &                      &                     & LSH & \textit{fail}     & \\
\hline
\multirow{6}{*}{T2} 
 & \multirow{3}{*}{IBM} & \multirow{3}{*}{20} & NH  & 151.0 $\pm$ 3.1   & \multirow{3}{*}{100} \\
 &                      &                     & OH  & 387.1 $\pm$ 18.8  & \\
 &                      &                     & LSH & 493.5 $\pm$ 18.0  & \\
\cline{2-6}
 & \multirow{3}{*}{NSE} & \multirow{3}{*}{20} & NH  & 131.0 $\pm$ 1.1   & \multirow{3}{*}{50} \\
 &                      &                     & OH  & \textit{fail}     & \\
 &                      &                     & LSH & \textit{fail}     & \\
\hline
\multirow{6}{*}{T3} 
 & \multirow{3}{*}{IBM} & \multirow{3}{*}{20} & NH  & 211.6 $\pm$ 2.4   & \multirow{3}{*}{100} \\
 &                      &                     & OH  & 511.5 $\pm$ 4.2   & \\
 &                      &                     & LSH & 545.2 $\pm$ 11.6  & \\
\cline{2-6}
 & \multirow{3}{*}{NSE} & \multirow{3}{*}{20} & NH  & 134.9 $\pm$ 0.3   & \multirow{3}{*}{50} \\
 &                      &                     & OH  & \textit{fail}     & \\
 &                      &                     & LSH & \textit{fail}     & \\
\end{tabular}

\vspace{2pt}
\footnotesize{
\textit{Notes:} 
NH – No hazards; 
LSH – Liquid spill hazards; 
OH – Object hazards; 
NSE – No skill execution.
}
\end{table}

The comparison of average task duration in T1 under \texttt{no hazard} conditions shows that using CIN results in only a 7.64\% increase in execution time over NSE, which we consider marginal balanced against the large safety enhancements. For tasks T2 and T3, IBM results in a 15.29\% and 56.76\% increase in execution time, respectively, compared to NSE. However, under \texttt{hazard} conditions, NSE consistently failed by: (1) collisions with obstructions, and (2) manipulation of labwares posing safety risks, leading to severe operational issues in all runs. Notably, the user would only become aware of these failures after the workflow had already broken down. These failures ultimately resulted in increased overall time due to the need for hazard clearance and/or workflow restart. These skills only take negligible amount of additional checking time during \texttt{no hazard} conditions and ultimately save huge amount of time and resources in \texttt{hazard} conditions.

%% file: Conclusion.tex
\section{Conclusion}

This paper presents PREVENT a system with two novel skills based on multi-modal perception and hierarchical decision-making for MRCs, aimed at ensuring safety during task execution. The proposed CIN and IBM skills leverage AI-driven vision and olfactory modalities to detect anomalies and enable context-aware decision-making. In cases involving hazards, the system requests intervention from human chemists before resuming the halted task. Experiments demonstrated that the proposed multi-modal strategies for both skills effectively avoid false positives and false negatives, consistently taking the expected actions when compared with other uni-modal and alternative multi-modal configurations. Quantitative experiments conducted on the identified modality configurations showed that, when tested individually, each modality failed in certain scenarios. However, the multi-modal configuration was able to detect these hazards reliably.  This improvement is attributed to the proposed multi-modal strategy, which compensates for the limitation of one modality by leveraging the strengths of the other.

While the current implementation is effective and shows promising results, the strategy remains task-centered rather than hazard-centered, which limits its generalization and requires further fine-tuning of the models in their present form before deployment in other tasks. Future work will focus on extending the system to handle more complex robotic manipulation lab tasks, such as solid addition, and on improving generalization by incorporating a hazard-centric approach.

%% file: main_elsarticle-template-num.bbl
\begin{thebibliography}{10}
\expandafter\ifx\csname url\endcsname\relax
  \def\url#1{\texttt{#1}}\fi
\expandafter\ifx\csname urlprefix\endcsname\relax\def\urlprefix{URL }\fi
\expandafter\ifx\csname href\endcsname\relax
  \def\href#1#2{#2} \def\path#1{#1}\fi

\bibitem{burger2020mobile}
B.~Burger, P.~M. Maffettone, V.~V. Gusev, C.~M. Aitchison, Y.~Bai, X.~Wang, X.~Li, B.~M. Alston, B.~Li, R.~Clowes, et~al., A mobile robotic chemist, Nature 583~(7815) (2020) 237--241.

\bibitem{lunt2024modular}
A.~M. Lunt, H.~Fakhruldeen, G.~Pizzuto, L.~Longley, A.~White, N.~Rankin, R.~Clowes, B.~Alston, L.~Gigli, G.~M. Day, et~al., Modular, multi-robot integration of laboratories: an autonomous workflow for solid-state chemistry, Chemical Science 15~(7) (2024) 2456--2463.

\bibitem{dai2024autonomous}
T.~Dai, S.~Vijayakrishnan, F.~T. Szczypi{\'n}ski, J.-F. Ayme, E.~Simaei, T.~Fellowes, R.~Clowes, L.~Kotopanov, C.~E. Shields, Z.~Zhou, et~al., Autonomous mobile robots for exploratory synthetic chemistry, Nature (2024) 1--8.

\bibitem{fakhruldeen2022archemist}
H.~Fakhruldeen, G.~Pizzuto, J.~Glowacki, A.~I. Cooper, Archemist: Autonomous robotic chemistry system architecture, in: 2022 International Conference on Robotics and Automation (ICRA), IEEE, 2022, pp. 6013--6019.

\bibitem{ghzouli2023}
R.~Ghzouli, T.~Berger, E.~B. Johnsen, A.~Wasowski, S.~Dragule, Behavior trees and state machines in robotics applications, IEEE Transactions on Software Engineering 49~(9) (2023) 4243--4267.
\newblock \href {https://doi.org/10.1109/TSE.2023.3269081} {\path{doi:10.1109/TSE.2023.3269081}}.

\bibitem{akkaladevi2025}
S.~C. Akkaladevi, M.~Ganglbauer, K.~Deshpande, S.~Ukleja, S.~C. Akkaladevi, A.~Pichler, Modular robot task execution framework for industrial pcb handling: A behavior tree approach, in: 2025 11th International Conference on Automation, Robotics, and Applications (ICARA), 2025, pp. 155--159.
\newblock \href {https://doi.org/10.1109/ICARA64554.2025.10977702} {\path{doi:10.1109/ICARA64554.2025.10977702}}.

\bibitem{martin2025towards}
F.~Mart{\'\i}n, E.~Soriano-Salvador, J.~M. Guerrero, G.~G. Muzquiz, J.~C. Manzanares, F.~J. Rodr{\'\i}guez, Towards a robotic intrusion prevention system: Combining security and safety in cognitive social robots, Robotics and Autonomous Systems 190 (2025) 104959.

\bibitem{Suddrey2022}
G.~Suddrey, B.~Talbot, F.~Maire, Learning and executing re-usable behaviour trees from natural language instruction, IEEE Robotics and Automation Letters 7~(4) (2022) 10643--10650.
\newblock \href {https://doi.org/10.1109/LRA.2022.3194681} {\path{doi:10.1109/LRA.2022.3194681}}.

\bibitem{Best2024}
G.~Best, R.~Garg, J.~Keller, G.~A. Hollinger, S.~Scherer, \href{https://doi.org/10.1177/02783649231203342}{Multi-robot, multi-sensor exploration of multifarious environments with full mission aerial autonomy}, The International Journal of Robotics Research 43~(4) (2024) 485--512.
\newblock \href {http://arxiv.org/abs/https://doi.org/10.1177/02783649231203342} {\path{arXiv:https://doi.org/10.1177/02783649231203342}}, \href {https://doi.org/10.1177/02783649231203342} {\path{doi:10.1177/02783649231203342}}.
\newline\urlprefix\url{https://doi.org/10.1177/02783649231203342}

\bibitem{Scherf2023}
L.~Scherf, A.~Schmidt, S.~Pal, D.~Koert, \href{https://www.frontiersin.org/journals/robotics-and-ai/articles/10.3389/frobt.2023.1152595}{Interactively learning behavior trees from imperfect human demonstrations}, Frontiers in Robotics and AI 10 (2023).
\newblock \href {https://doi.org/10.3389/frobt.2023.1152595} {\path{doi:10.3389/frobt.2023.1152595}}.
\newline\urlprefix\url{https://www.frontiersin.org/journals/robotics-and-ai/articles/10.3389/frobt.2023.1152595}

\bibitem{kato2018}
S.~Kato, S.~Tokunaga, Y.~Maruyama, S.~Maeda, M.~Hirabayashi, Y.~Kitsukawa, A.~Monrroy, T.~Ando, Y.~Fujii, T.~Azumi, Autoware on board: Enabling autonomous vehicles with embedded systems, in: 2018 ACM/IEEE 9th International Conference on Cyber-Physical Systems (ICCPS), 2018, pp. 287--296.
\newblock \href {https://doi.org/10.1109/ICCPS.2018.00035} {\path{doi:10.1109/ICCPS.2018.00035}}.

\bibitem{akkaladevi2024}
S.~C. Akkaladevi, M.~Propst, K.~Deshpande, M.~Hofmann, A.~Pichler, Towards a behavior tree based robotic skill execution framework for human robot collaboration in industrial assembly, in: 2024 10th International Conference on Automation, Robotics and Applications (ICARA), 2024, pp. 18--22.
\newblock \href {https://doi.org/10.1109/ICARA60736.2024.10553029} {\path{doi:10.1109/ICARA60736.2024.10553029}}.

\bibitem{Tanaka2023}
Y.~Tanaka, S.~Katsura, Task switching model for acceleration control of multi-dof manipulator using behavior trees, in: IECON 2023- 49th Annual Conference of the IEEE Industrial Electronics Society, 2023, pp. 1--6.
\newblock \href {https://doi.org/10.1109/IECON51785.2023.10311686} {\path{doi:10.1109/IECON51785.2023.10311686}}.

\bibitem{leggett2012identifying}
D.~J. Leggett, Identifying hazards in the chemical research laboratory, Process Safety Progress 31~(4) (2012) 393--397.

\bibitem{menard2020review}
A.~D. M{\'e}nard, J.~F. Trant, A review and critique of academic lab safety research, Nature chemistry 12~(1) (2020) 17--25.

\bibitem{kong2018li}
L.~Kong, C.~Li, J.~Jiang, M.~G. Pecht, Li-ion battery fire hazards and safety strategies, Energies 11~(9) (2018) 2191.

\bibitem{cooper2025lira}
A.~Cooper, Z.~Zhou, S.~Veeramani, F.~Galeano, H.~Fakhruldeen, Lira: Localization, inspection, and reasoning module for autonomous workflows in self-driving labs (2025).

\bibitem{liu2012floyd}
H.~Liu, N.~Stoll, S.~Junginger, K.~Thurow, A floyd-genetic algorithm based path planning system for mobile robots in laboratory automation, in: 2012 IEEE International Conference on Robotics and Biomimetics (ROBIO), IEEE, 2012, pp. 1550--1555.

\bibitem{liu2013mobile}
H.~Liu, N.~Stoll, S.~Junginger, K.~Thurow, Mobile robot for life science automation, International Journal of Advanced Robotic Systems 10~(7) (2013) 288.

\bibitem{jiang2023autonomous}
Y.~Jiang, H.~Fakhruldeen, G.~Pizzuto, L.~Longley, A.~He, T.~Dai, R.~Clowes, N.~Rankin, A.~I. Cooper, Autonomous biomimetic solid dispensing using a dual-arm robotic manipulator, Digital Discovery 2~(6) (2023) 1733--1744.

\bibitem{munguia2023affordance}
F.~Munguia-Galeano, S.~Veeramani, J.~D. Hern{\'a}ndez, Q.~Wen, Z.~Ji, Affordance-based human--robot interaction with reinforcement learning, IEEE Access 11 (2023) 31282--31292.

\bibitem{song2024vlm}
D.~Song, J.~Liang, A.~Payandeh, A.~H. Raj, X.~Xiao, D.~Manocha, Vlm-social-nav: Socially aware robot navigation through scoring using vision-language models, IEEE Robotics and Automation Letters (2024).

\bibitem{munguia2025chemist}
F.~Munguia-Galeano, Z.~Zhou, S.~Veeramani, H.~Fakhruldeen, L.~Longley, R.~Clowes, A.~I. Cooper, Chemist eye: A visual language model-powered system for safety monitoring and robot decision-making in self-driving laboratories, arXiv preprint arXiv:2508.05148 (2025).

\bibitem{pizzuto2024accelerating}
G.~Pizzuto, H.~Wang, H.~Fakhruldeen, B.~Peng, K.~S. Luck, A.~I. Cooper, Accelerating laboratory automation through robot skill learning for sample scraping, in: 2024 IEEE 20th International Conference on Automation Science and Engineering (CASE), IEEE, 2024, pp. 2103--2110.

\bibitem{song2024multiagent}
T.~Song, M.~Luo, X.~Zhang, L.~Chen, Y.~Huang, J.~Cao, Q.~Zhu, D.~Liu, B.~Zhang, G.~Zou, et~al., A multiagent-driven robotic ai chemist enabling autonomous chemical research on demand, Journal of the American Chemical Society (2024).

\bibitem{al2024ambient}
M.~F.~R. Al-Okby, S.~Junginger, T.~Roddelkopf, J.~Huang, K.~Thurow, Ambient monitoring portable sensor node for robot-based applications, Sensors 24~(4) (2024) 1295.

\bibitem{DARVISH2025101897}
K.~Darvish, M.~Skreta, Y.~Zhao, N.~Yoshikawa, S.~Som, M.~Bogdanovic, Y.~Cao, H.~Hao, H.~Xu, A.~Aspuru-Guzik, A.~Garg, F.~Shkurti, \href{https://www.sciencedirect.com/science/article/pii/S2590238524005423}{Organa: A robotic assistant for automated chemistry experimentation and characterization}, Matter 8~(2) (2025) 101897.
\newblock \href {https://doi.org/https://doi.org/10.1016/j.matt.2024.10.015} {\path{doi:https://doi.org/10.1016/j.matt.2024.10.015}}.
\newline\urlprefix\url{https://www.sciencedirect.com/science/article/pii/S2590238524005423}

\bibitem{fakhruldeen2025multimodal}
H.~Fakhruldeen, A.~R. Nambiar, S.~Veeramani, B.~V. Tailor, H.~B. Juneghani, G.~Pizzuto, A.~I. Cooper, Multimodal behaviour trees for robotic laboratory task automation, in: 2025 IEEE International Conference on Robotics and Automation (ICRA), 2025, pp. 15872--15878.
\newblock \href {https://doi.org/10.1109/ICRA55743.2025.11128408} {\path{doi:10.1109/ICRA55743.2025.11128408}}.

\bibitem{Zhou2025VLM}
Z.~Zhou, S.~Veeramani, H.~Fakhruldeen, S.~Uyanik, A.~I. Cooper, Genco: A dual vlm generate-correct framework for adaptive peg-in-hole robotics, in: 2025 IEEE International Conference on Robotics and Automation (ICRA), 2025, pp. 16744--16751.
\newblock \href {https://doi.org/10.1109/ICRA55743.2025.11128409} {\path{doi:10.1109/ICRA55743.2025.11128409}}.

\bibitem{pizzuto2022solis}
G.~Pizzuto, J.~De~Berardinis, L.~Longley, H.~Fakhruldeen, A.~I. Cooper, Solis: Autonomous solubility screening using deep neural networks, in: 2022 International Joint Conference on Neural Networks (IJCNN), IEEE, 2022, pp. 1--7.

\bibitem{radford2021learning}
A.~Radford, J.~W. Kim, C.~Hallacy, A.~Ramesh, G.~Goh, S.~Agarwal, G.~Sastry, A.~Askell, P.~Mishkin, J.~Clark, et~al., Learning transferable visual models from natural language supervision, in: International conference on machine learning, PmLR, 2021, pp. 8748--8763.

\bibitem{dorbala2022clip}
V.~S. Dorbala, G.~Sigurdsson, R.~Piramuthu, J.~Thomason, G.~S. Sukhatme, Clip-nav: Using clip for zero-shot vision-and-language navigation, arXiv preprint arXiv:2211.16649 (2022).

\bibitem{ZHANG2025105210}
X.~Zhang, H.~Ling, X.~Huang, Q.~Jin, J.~Hu, \href{https://www.sciencedirect.com/science/article/pii/S0921889025003070}{Ovgrasp: Target-oriented open-vocabulary robotic grasping in clutter}, Robotics and Autonomous Systems (2025) 105210\href {https://doi.org/https://doi.org/10.1016/j.robot.2025.105210} {\path{doi:https://doi.org/10.1016/j.robot.2025.105210}}.
\newline\urlprefix\url{https://www.sciencedirect.com/science/article/pii/S0921889025003070}

\bibitem{muttaqien2025attention}
M.~A. Muttaqien, T.~Motoda, R.~Hanai, D.~Yukiyasu, Attention-guided integration of clip and sam for precise object masking in robotic manipulation, in: 2025 IEEE/SICE International Symposium on System Integration (SII), IEEE, 2025, pp. 1213--1220.

\bibitem{gao2024physically}
J.~Gao, B.~Sarkar, F.~Xia, T.~Xiao, J.~Wu, B.~Ichter, A.~Majumdar, D.~Sadigh, Physically grounded vision-language models for robotic manipulation, in: 2024 IEEE International Conference on Robotics and Automation (ICRA), IEEE, 2024, pp. 12462--12469.

\end{thebibliography}
